\documentclass{article}
\usepackage{spconf,amsmath,graphicx, setspace, booktabs}
\usepackage{lipsum}
\usepackage{tabularx}
\usepackage{multirow}

\title{Wavelet-Guided Acceleration of Text Inversion \\in Diffusion-Based Image Editing}
\name{Gwanhyeong Koo, Sunjae Yoon, Chang D. Yoo\protect\textsuperscript{\dag} \thanks{\protect\textsuperscript{\dag}Corresponding author}  
\thanks{This work was partly supported by Center for Applied Research in Artificial Intelligence (CARAI) grant funded by DAPA and ADD (UD230017TD) and partly supported by the Institute for Information \& communications Technology Promotion (IITP) grant funded by the Korea government (MSIT) (No. 2021-0-01381, Development of Causal AI through Video Understanding and Reinforcement Learning, and Its Applications to Real Environments).}}

\address{Korea Advanced Institute of Science and Technology (KAIST), Daejeon, Republic of Korea \\
}

\begin{document}

\maketitle

\begin{abstract}
In the field of image editing, Null-text Inversion (NTI) enables fine-grained editing while preserving the structure of the original image by optimizing null embeddings during the DDIM sampling process. However, the NTI process is time-consuming, taking more than two minutes per image. To address this, we introduce an innovative method that maintains the principles of the NTI while accelerating the image editing process. We propose the WaveOpt-Estimator, which determines the text optimization endpoint based on frequency characteristics. Utilizing wavelet transform analysis to identify the image's frequency characteristics, we can limit text optimization to specific timesteps during the DDIM sampling process. By adopting the Negative-Prompt Inversion (NPI) concept, a target prompt representing the original image serves as the initial text value for optimization. This approach maintains performance comparable to NTI while reducing the average editing time by over 80\% compared to the NTI method. Our method presents a promising approach for efficient, high-quality image editing based on diffusion models.

\end{abstract}

\section{Introduction}
\label{sec:intro}

In recent years, diffusion \cite{DDPM} has achieved remarkable progress in image generation, outperforming the capabilities and addressing the challenges faced by previous Generative Adversarial Network (GAN) \cite{gan_2014, stackgan, gan_overview, gan_2019} approaches. While many recent diffusion-based models \cite{ldm, dalle2, imagen, glide, stylediffusion} have demonstrated proficiency in text-to-image generation tasks, they have consistently faced challenges in image editing, especially in maintaining the original structure and details. Addressing this issue, Prompt-to-Prompt \cite{prompt_to_prompt} introduced a technique that leverages cross-attention guidance to achieve more refined edits. Although this method marked an improvement, it struggled to preserve the original image's fine details and subtle features. To overcome this limitation, Null-text Inversion (NTI) \cite{nti} significantly advanced the field. By optimizing null embeddings during the DDIM \cite{ddim} sampling, NTI effectively enabled subtle edits that were closely aligned with the original image's attributes.

However, a significant drawback of null-text optimization is its time-consuming nature. In practice, this process can take upwards of two minutes for a single image. Efforts are underway to accelerate null-text optimization, with Negative-Prompt Inversion (NPI) \cite{npi} being a notable attempt. By using the original text prompt embedding instead of the null-text, the need for optimization is bypassed, allowing it to operate at times comparable to the DDIM Inversion \cite{ddim, diffusion_beat_gans}. Nonetheless, the average reconstruction quality falls short compared to null-text inversion.

In this paper, we propose a novel approach to maximize the efficiency of NTI while also enhancing processing speed. We discover that optimizing only up to a specific timestep and applying the last null-text embedding vector to all subsequent timesteps consistently results in outstanding performance. This observation prompted us to explore methods for dynamically determining the optimal stopping point for different images. Building on this idea, we analyze the image frequency components using wavelet transform. Our analysis reveals that the time required for text optimization varies based on the frequency characteristics of the image. Inspired by these findings, we introduce the WaveOpt-Estimator (Wavelet-based Optimization Estimator). Employing this estimator, we can identify a specific timestep to cease optimization depending on the image's frequency characteristics, presenting a novel approach to enhancing efficiency in DDIM Inversion-based image editing \cite{prompt_to_prompt, diffedit, plug_and_play, masactrl}. 

\begin{figure}[htb!]
  \centering
  \includegraphics[width=\linewidth]{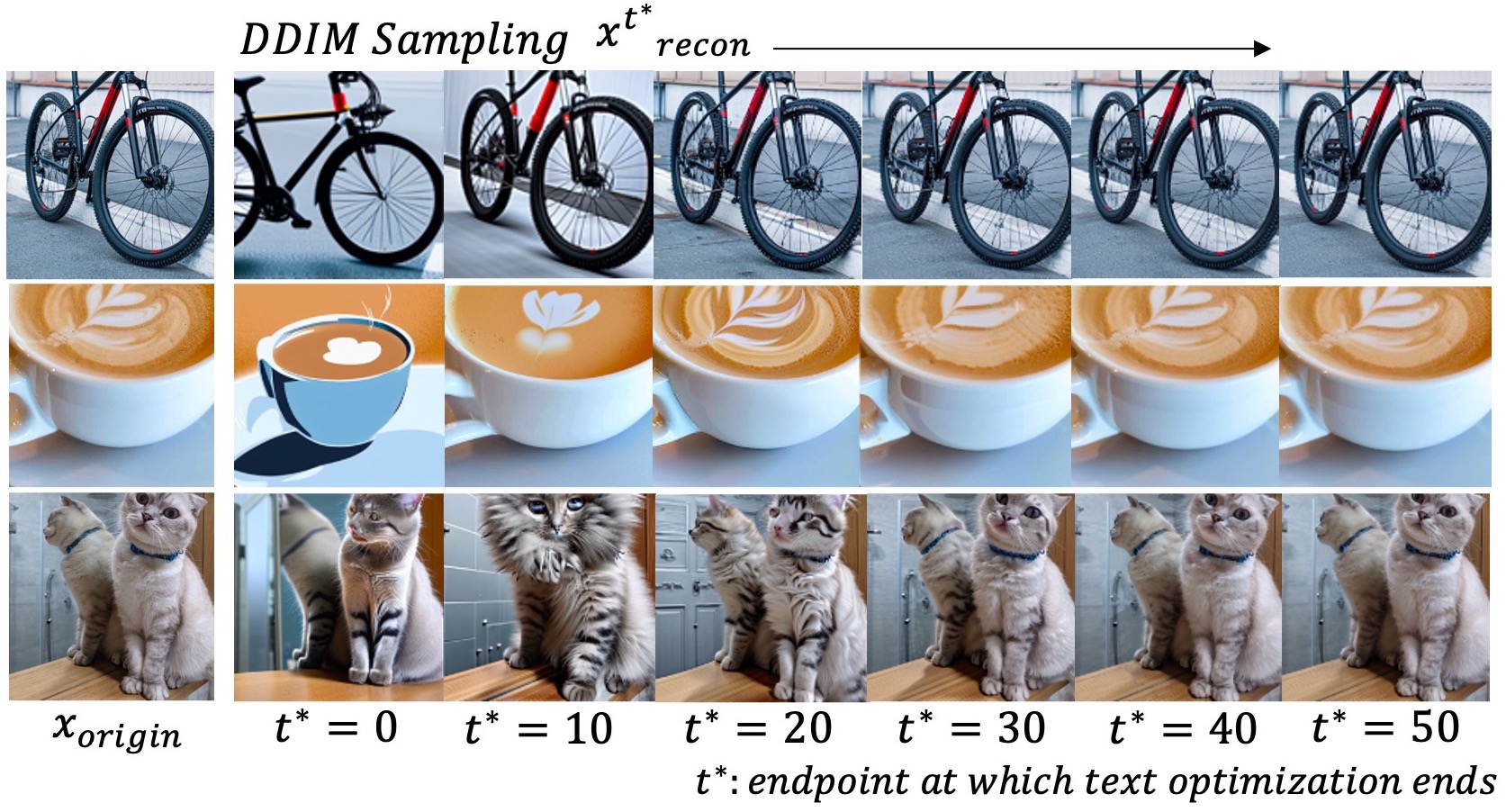}
   \caption{Results of DDIM Sampling According to $t^*$}
   \label{fig:ddim_sampling}
\end{figure}

Through extensive experiments, the results show that optimizing only up to certain timesteps achieves satisfactory image quality. Moreover, by leveraging the features of NPI, we set the initial null-text as a target text prompt that accurately represents the original image, speeding up the image reconstruction process while keeping pace with the accuracy of prior methods. Remarkably, most images completed their optimization within 20 timesteps, leading to over an 80\% reduction in processing time compared to NTI methods and surpassing the accuracy of NPI. A more in-depth analysis of our findings will be presented in Section 2.





\section{Preliminary Analysis}
\label{sec:pagestyle}

\subsection{Null-text Optimization Analysis}
\label{ssec:subhead}
In this section, we investigate the impact on performance when Null-Text Inversion (NTI) is applied not throughout all timesteps but only up to a specific timestep. To set the stage for our analysis, we first introduce some terminology related to the basic DDIM inversion process and null-text optimization. Let $x_{ori}$ represent the original image. We denote the total number of timesteps for DDIM inversion and sampling as $T$.

\vspace{-0.5cm}
\begin{align}
& z_{0} = E(x_{ori})  \\
& z_{T} = \text{DDIM}_{\text{Inv}}(z_{0}, p_{src})  \\
& \hat{z_{0}} = \text{DDIM}(z_{T}, p_{src}, p_{edit}, {\left\{{{\phi_t}} \right\}}^{T}_{t=1}  )   \\
& x_{edit} = D(\hat{z_{0}})  
\end{align}

When the image editing is performed, $x_{ori}$ is passed through an Encoder $E$ to become the latent $z_{0}$ (1), and it is used in the DDIM inversion process along with the source prompt $p_{src}$, which represents the original image, to generate a noisy latent $z_{T}$ (2). The UNet then takes $z_{T}$, the source prompt $p_{src}$, the editing prompt $p_{edit}$, and the pre-optimized null-text embedding values ${\left\{{{\phi_t}} \right\}}^{T}_{t=1}$ to perform denoising, producing a modified latent $\hat{z{0}}$ that retains characteristics of the original image while incorporating the modifications specified by $p_{edit}$ (3). The modified $\hat{z_{0}}$ is then passed through the Decoder $D$ to produce the image, which combines the original structure with the intended edits (4).

\vspace{-0.5cm}
\begin{align}
& \phi^{copy}_{i} = \phi_{t^*} \quad for \quad i \: \in \: [t^*+1, T] \\
& {\left\{{{\phi^{copy}_t}} \right\}}^{T}_{t=1} = [\phi_{1}, \phi_{2}, ..., \phi_{t^*}, \phi^{copy}_{t^*}, ..., \phi^{copy}_{t^*}]
\end{align}

\begin{figure}[htb!]
  \includegraphics[width=\linewidth]{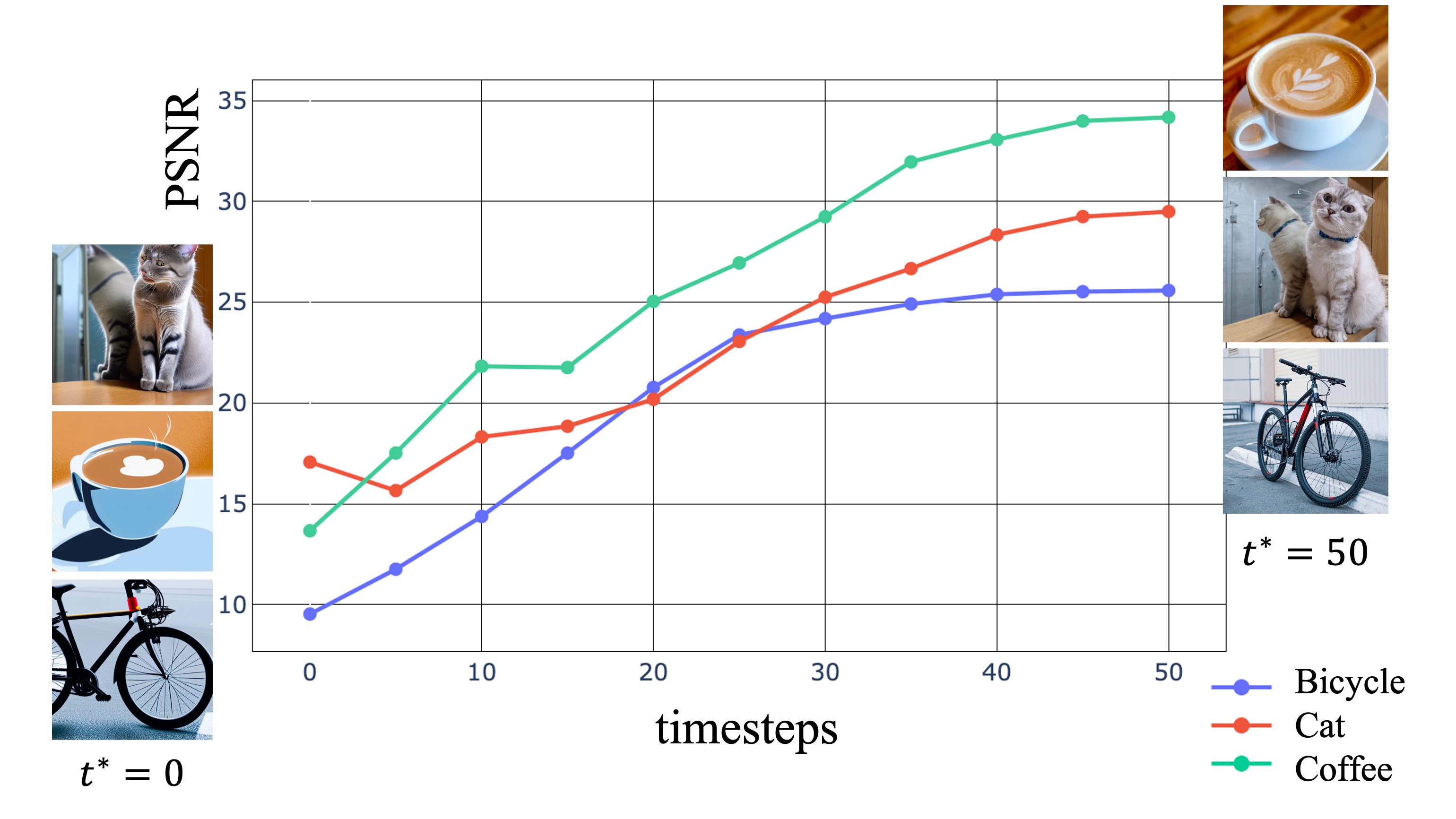}
   \vspace{-0.5cm}
   \caption{Results of PSNR between $x_{ori}$ and $x^{i^*}_{recon}$}
   \label{fig:psnr}
\end{figure}

We introduce the concept of ${\left\{{{\phi^{copy}_t}} \right\}}^{T}_{t=1}$ (6), a sequence of null-text embedding vectors obtained by applying NTI up to a specific timestep $t^*$, and subsequently duplicating the last null-text $\phi_{t^*}$ to expand the remaining timesteps up to $T$ (5). 

\begin{align}
& \hat{z^{*}_{0}} = \text{DDIM}(z_{T}, p_{src}, {\left\{{{\phi^{copy}_{t}}} \right\}}^{T}_{t=1} \ with \ t^* ) \\
& x^{t*}_{recon} = D(\hat{z^{*}_{0}}) \\
& X_{recon} = {\left\{{{x^{t^*}_{recon}}} \right\}}^{T}_{t^{*}=1}
\end{align}

The reconstructed image obtained by passing the latent $\hat{z^{*}_{0}}$ (7) through the decoder $D$ is denoted as $x^{t^*}_{recon}$ (8). In here, the latent $\hat{z^{*}_{0}}$ used for generating $x^{t^*}_{recon}$ is derived from employing DDIM sampling with ${\left\{{{\phi^{copy}_t}} \right\}}^{T}_{t=1}$, where NPI is contained to a specific timestep $t^*$.  In practice, we set $T$ at 50 and conduct experiments with $t^*$ ranging from 0 to $T$, incrementing by strides of 5. The set of $x^{t^*}_{recon}$ obtained at different $t^*$ is denoted as $X_{recon}$ (9) and can be observed in Figure \ref{fig:ddim_sampling}. To assess the similarity between the reconstructed $x^{t^*}_{recon}$ and the original image $x_{ori}$, the PSNR metric is utilized, and the outcomes are shown in Figure \ref{fig:psnr}. It is found that the PSNR value converges differently for each image as the timestep progresses, and after a certain timestep, the incremental gain in PSNR becomes marginal. 

For instance, in Figure \ref{fig:psnr}, the image of a bicycle reaches a PSNR value of 24.9 by the 25th timestep, accounting for 97.4\% of its PSNR value at the 50th timestep, which is 25.65. On the other hand, the image of coffee attains a PSNR value of 33.7 at the 35th timestep, which is 97.1\% of its PSNR value at the 50th timestep, 34.36. In this context, the ratio of the PSNR at a specific timestep $t^*$ to that at the 50th timestep $T$ is defined as the PSNR ratio. 

Empirically, it is observed that the differences between the original and the reconstructed images are negligible when the PSNR ratio is above 0.9. We denote timestep $t^*$ as the endpoint where the PSNR ratio first exceeds 0.9. This study posits that the endpoint is determined by unique image characteristics, thereby serving as motivation for further study in the next section.

\subsection{Frequency Analysis}
\label{ssec:subhead}
\begin{figure}[htb!]
  \centerline{\includegraphics[width=\linewidth]{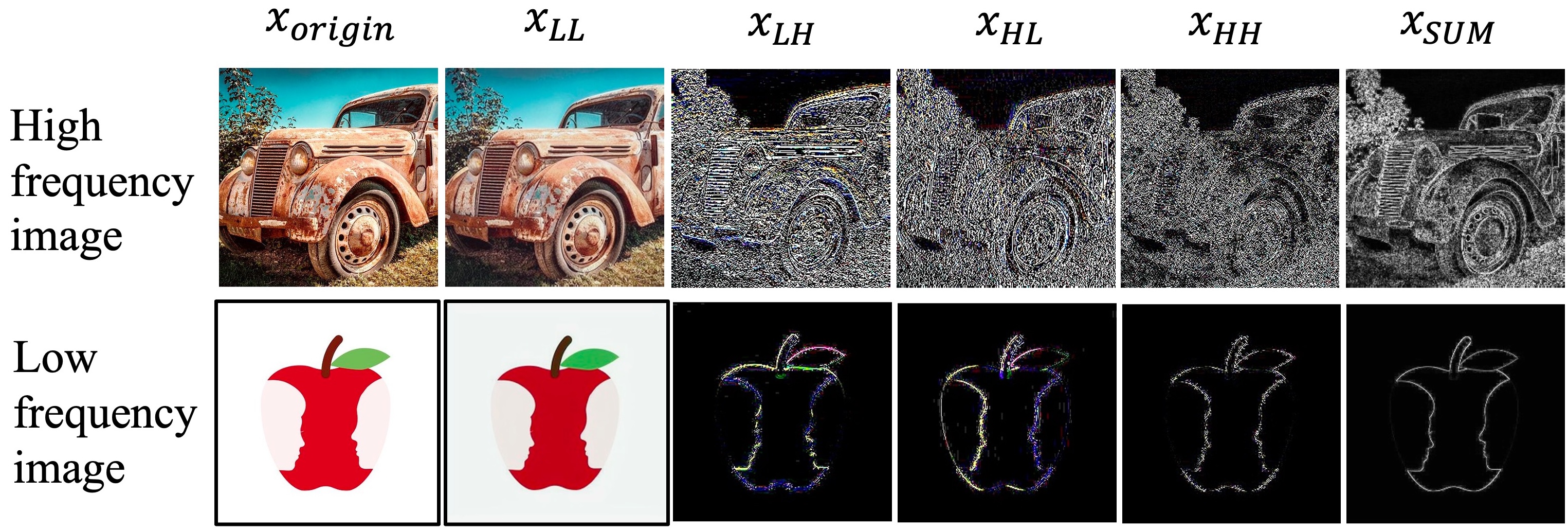}}
   \caption{Wavelet subbands and its energy}
   \label{fig:wavelet}
\end{figure}

To explore how quickly the PSNR ratio converges based on the inherent characteristics of the image, we use the Discrete Wavelet Transform (DWT). DWT decomposes an image $x$ into frequency subbands $x_{LL}, x_{LH}, x_{HL}, x_{HH}$ using a set of low and high-pass filters. For a more precise analysis of high-frequency components, we compute the average of the $ x_{LH}, x_{HL}$, and $x_{HH}$ subbands, resulting in $x_{SUM}$. Further, we enhance $x_{SUM}$ using adaptive histogram equalization to accentuate high-frequency details. We focus our analysis on the $x_{LL}, x_{SUM}$ subbands and calculate their energy, which is the square of the signal (10-11). These results are presented in Figure \ref{fig:wavelet} and Table \ref{tab:frequency_psnr}, where the $t^*$ denotes the endpoint.

\vspace{-0.1cm}

\vspace{-0.5cm}
\begin{table}[htb!]
  \centering
  \footnotesize
  \caption{PSNR ratio analysis according to the image type}
  \begin{tabularx}{\linewidth}{c|c|c|c|c}
    \toprule
    \multicolumn{1}{c|}{Image type} & \multicolumn{1}{c|}{$E(x_{LL})$} & \multicolumn{1}{c|}{$E(x_{SUM})$} & \multicolumn{1}{c|}{endpoint $t^*$} & \multicolumn{1}{c}{PSNR ratio} \\
    \midrule
    High frequency &16756 &8714 &5  &0.943\\
    Low frequency  &49642 &373 &25  &0.955\\
    \bottomrule
  \end{tabularx}
  \label{tab:frequency_psnr}
\end{table}

The experimental results reveal that images with prominent high-frequency components tend to exhibit lower energy in the $x_{LL}$ and higher energy in the $x_{SUM}$. Conversely, images rich in low-frequency components show higher energy in the $x_{LL}$ and lower energy in the $x_{SUM}$.
Notably, we discovered that images with lower energy in $x_{LL}$ and higher energy in $x_{SUM}$ tend to reach the endpoint $t^*$ at lower timesteps during DDIM sampling. This observation suggests that the UNet's DDIM sampling requires more time to form low-frequency elements, while forming high-frequency elements is comparatively faster, impacting the pace of image reconstruction. This finding indicates that the frequency components of an image affect the speed of convergence during DDIM sampling and the time required for text optimization.

\vspace{-0.1cm}
{\footnotesize
\begin{align}
& \text{Energy}(x_{LL}) = \sum_{i, j} x_{LL}[i, j]^{2} \\ 
& \text{Energy}(x_{SUM}) = \sum_{i, j} \left( \frac{x_{LH}[i, j] + x_{HL}[i, j] + x_{HH}[i, j]}{3} \right)^2
\end{align}
}


\section{METHOD}
\label{sec:pagestyle}
In the analysis presented in Section 2, a dependency is observed between an image's inherent frequency components and endpoint $t^*$. Inspired by these findings, we propose WaveOpt-Estimator, a model to estimate the endpoint $t^*$ based on the frequency characteristics of an image. Using predicted endpoint $t^*$, ${\left\{{{\phi^{copy}_t}} \right\}}^{T}_{t=1}$ is generated, significantly reducing the time required for optimization compared to the previous approach of performing optimization across all timesteps in NTI. Additionally, rather than using null-text as an initial value, we leverage the advantages of NPI to set the initial text value to a target prompt that describes the original image. 

\begin{figure}[t]
  \centering
  \includegraphics[width=\linewidth]{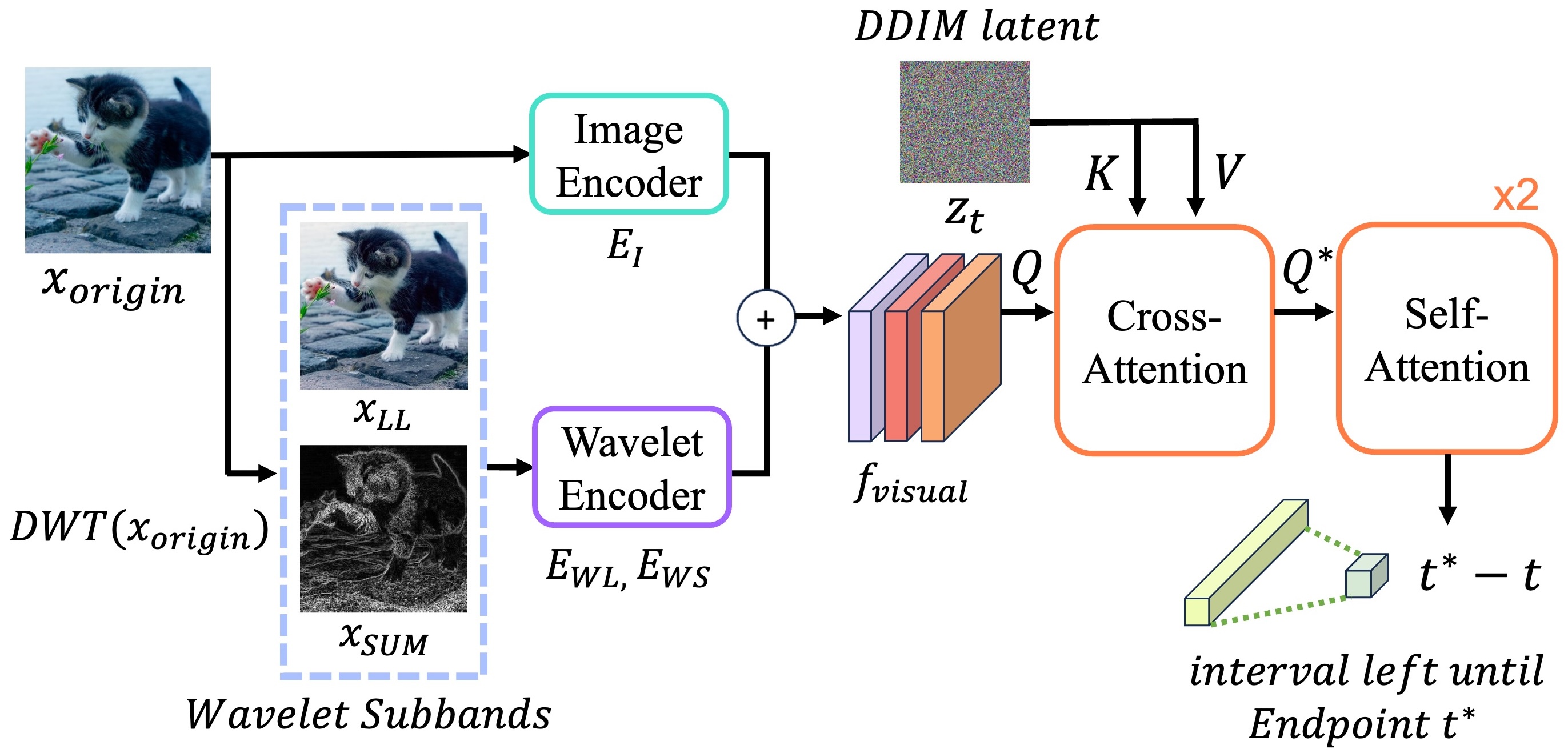} 
  \vspace{-0.8cm}
   \caption{Overview of our proposed WaveOpt-Estimator model}
   \label{fig:model}
\end{figure}

\vspace{-0.2cm}
\subsection{Model Architecture}
\label{ssec:subhead}

The WaveOpt-Estimator comprises encoders for processing the original image and wavelet subbands, along with cross-attention and self-attention modules. The $x_{ori}$ is decomposed into $x_{LL}$, $x_{SUM}$ subbands through Discrete Wavelet Transform (DWT). Subsequently, $x_{ori}$ is encoded via Image Encoder $E_{I}$ (13), while $x_{LL}$ and $x_{SUM}$ are separately encoded using a Wavelet Encoder $E_{WL}$ and $E_{WS}$ (14). ResNet-50 is employed for both Image and Wavelet Encoder. These encoded features are concatenated to form a unified visual feature $f_{visual}$ (15). 
\vspace{-0.5cm}

\begin{spacing}{1}
\begin{align}
& f_{origin} = E_I(x_{ori})\\
& f_{LL} = E_{WL}(x_{LL}), \; f_{SUM} = E_{WS}(x_{SUM}) \\
& f_{visual} = concat(f_{ori}, f_{LL}, f_{SUM}) \\ 
& Q^* = CA(f_{visual}, z_{t}) \quad for \quad t \: \in \: [0, t^*]
\end{align}
\end{spacing}

To train the multimodal relationship between visual features and the DDIM latent variables, cross-attention is applied. In the cross-attention module, the visual feature $f_{visual}$ serves as the Query, and $z_{t}$, stored during the DDIM inversion process, serves as both the Key and Value. For a given input image $x_{ori}$, $z_{t}$ exists for a total of $t^*$ times, from $t=0$ to $t=t^*$. Following the cross-attention (CA) between the visual feature and $z_t$, the resultant $Q^*$ serves as the input for subsequent self-attention (16). After two self-attention, the output tensor undergoes a fully connected layer for regression. The regressed value represents the duration from the current timestep $t$ to the endpoint $t^{*}$, where optimization should be limited. 


\subsection{Loss Function}
\label{ssec:subhead}

The WaveOpt-Estimator is devised to predict the endpoint $t^*$ using the original image $x_{ori}$, its frequency components ${x_{LL}}$ and ${x_{SUM}}$, and the DDIM latent variable $z_{t}$. To achieve this, an L2 loss is applied between the predicted endpoint $t^{*}_{pred}$ and the ground truth $t^{*}_{gt}$ (17). Additionally, a hinge loss based on PSNR is utilized between $x_{ori}$ and the image derived from $z_{t}$ (18). We define $PSNR_t$ as the PSNR between the $x_{ori}$ and the reconstructed image from $z_{t}$, and $PSNR_t^{*}$ for the one from the endpoint $z_t$. This loss captures the discrepancy in PSNR between specific timestep $t$ and the endpoint $t^*$. Experiments are conducted by applying weight factors $\alpha_1$, and $\alpha_2$ to the two loss terms (19), setting both factors to 0.5.

\vspace{-0.5cm}
\begin{align}
& \mathcal{L}_{L2, t} = ||(t^{*}_{pred} - t) - (t^{*}_{gt} - t)||^{2}_{2} \\
& \mathcal{L}_{hinge, t} = max(0, PSNR_{t^*} - PSNR_{t}) \\
& \mathcal{L}_{t} = \alpha_1 * \mathcal{L}_{L2, t} + \alpha_2 * \mathcal{L}_{hinge, t}
\end{align}


\section{EXPERIMENTS}
\label{sec:typestyle}

\subsection{Dataset and Evaluation Metric}
\label{ssec:subhead}

We assemble a dataset consisting of 50 images from various papers \cite{stylediffusion, nti, imagic, proxinpi}. For each image, we generate $(x_{ori}, x_{LL}, x_{SUM}, z_{t})$ pairs for every timestep $t$ from $t=0$ to $t=t^*$ during DDIM inversion process. Each image produces $t^*$ pairs, resulting in  $30 \times t^*$ training pairs and $10 \times t^*$ pairs for both validation and test datasets. In practice, our training data consists of 750 samples, with the validation and test data containing 250 samples each. Training is conducted on a Quadro RTX 8000 GPU, and for the evaluation metric, we adopt the MAE(Mean Absolute Error) metric. The MAE quantify the difference between the predicted value $t^{*}_{pred} - t$ and the ground truth value $t^{*}_{gt} - t$.

\vspace{-0.5cm}

\begin{table}[htb!]
  \centering
  \footnotesize
  \caption{Performance Metrics for WaveOpt-Estimator}
  \begin{tabularx}{\linewidth}{@{}>{\hsize=0.6\hsize\centering\arraybackslash}X|
                                               >{\hsize=0.8\hsize\centering\arraybackslash}X|
                                               >{\hsize=0.8\hsize\centering\arraybackslash}X|
                                               >{\hsize=0.8\hsize\centering\arraybackslash}X|
                                               >{\hsize=0.8\hsize\centering\arraybackslash}X@{}}
    \toprule
    \multirow{2}{*}{Epoch} & \multicolumn{2}{c|}{Training} & \multicolumn{2}{c}{Test} \\     
       & loss & MAE $\downarrow$ & loss & MAE $\downarrow$ \\
    \midrule
     10   & 3.26  & 5.31  & 5.24  & 7.75\\
     30   & 0.53  & 2.1   & 3.15  & 5.4\\
     50   & 0.32  & 1.5   & 2.02    & 2.9\\
    \bottomrule
  \end{tabularx}
  \label{tab:comparison}
\end{table}

\vspace{-0.5cm}

\subsection{Experimental Results}
\label{ssec:subhead}
\begin{figure}[htb!]
  \centering
  \includegraphics[width=\linewidth]{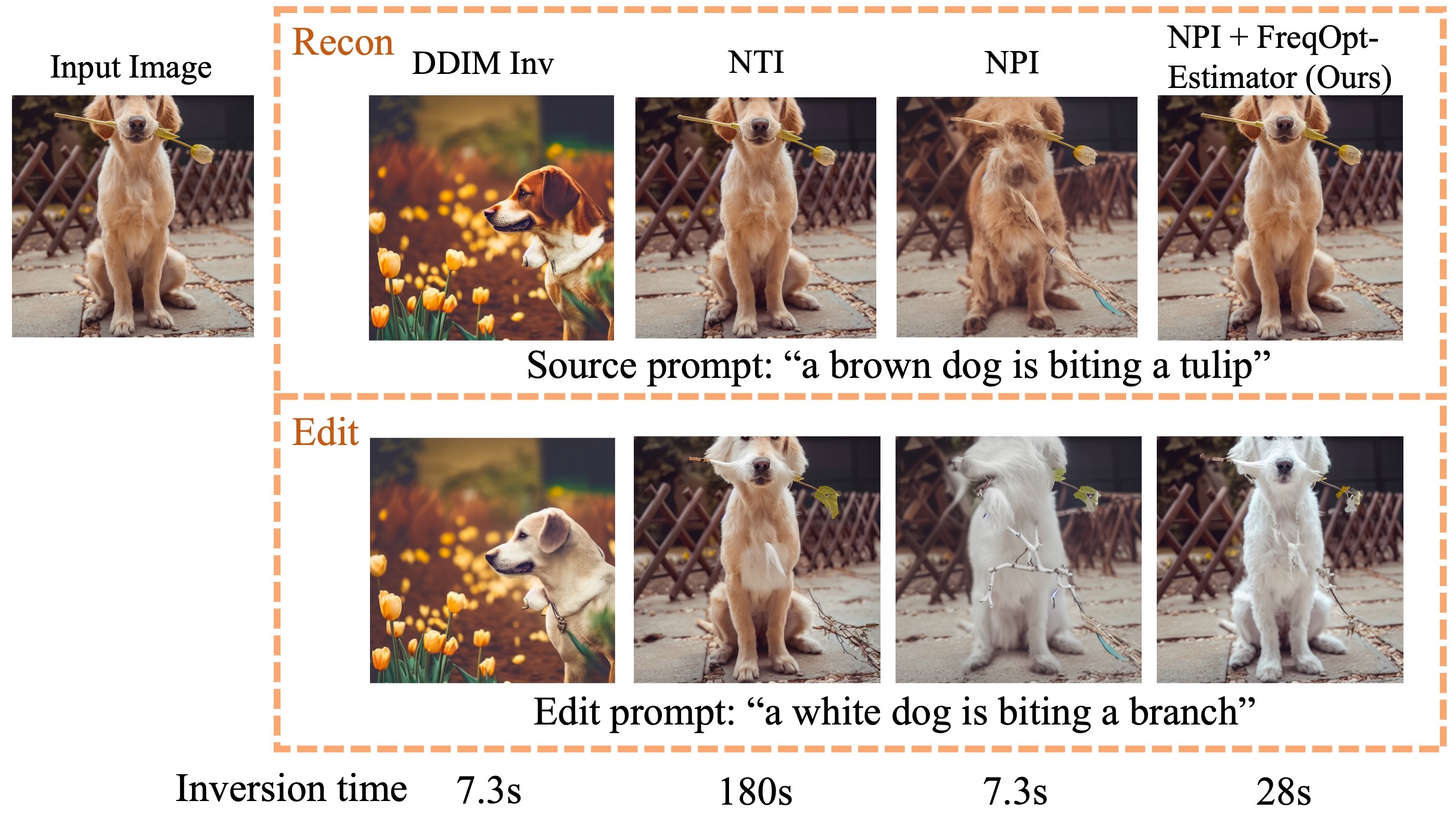} 
  \vspace{-0.8cm}
   \caption{Comparison of reconstruction and edit results with other methods}
   \label{fig:model}
\end{figure}

\vspace{-0.3cm}

Based on the configurations provided in Section 4, when training the WaveOpt-Estimator, we observe that it converged to a Test MAE of 2.9 by Epoch 50, as detailed in Table \ref{tab:comparison}. This outcome reveals that the endpoint predicted by the WaveOpt-Estimator, $t^*_{pred}$, has an error margin of 3 timesteps when compared to the actual $t^*_{gt}$. With this information, we can effectively determine an appropriate stopping point for text optimization during the DDIM inversion process. Given the MAE of 2.9, we would continue for an additional 3 timesteps beyond the identified stopping point to ensure the best results.

Subsequently, we evaluate the performance of the WaveOpt-Estimator, as can be seen in Table \ref{tab:comparison2}. We set the Prompt-to-Prompt as our baseline and compared both NTI and NPI with our proposed method. In this context, the average time denotes the duration taken for DDIM Inversion, inclusive of text optimization, and does not incorporate the time spent on sampling. The NTI approach took approximately 180 seconds per image but showcased superior performance in terms of the PSNR ratio and SSIM. On the other hand, while NPI exhibited a faster execution time, similar to the DDIM inversion method at around 7.3 seconds, severe distortions occurred during the reconstruction of some images, as illustrated in Figure 5. By applying the WaveOpt-Estimator to both NTI and NPI, we not only achieved a PSNR ratio exceeding 0.9 but also reduced the processing time from 180 seconds to 28 seconds, marking an approximately 80\% improvement. The time taken by the WaveOpt-Estimator to predict $t^*_{pred}$ is within 5 seconds and is included in the average processing time of Table \ref{tab:comparison2}. 

\vspace{-0.2cm}
\begin{table}[ht]
  \centering
  \footnotesize
  \caption{Image Editing Quality and Speed Comparison}
  \begin{tabularx}{\linewidth}{@{}l|c|c|c@{}}
    \toprule
    \multicolumn{1}{l|}{Method} & \multicolumn{1}{c|}{PSNR ratio $\uparrow$} & \multicolumn{1}{c|}{SSIM $\uparrow$} & \multicolumn{1}{c}{Average time (s)}  \\
    \midrule
    DDIM inversion with CFG &0.68 &0.61 &7.3  \\
    NTI  &1 &0.92   &180  \\
    NPI  &0.82 &0.73   &7.3  \\
    NTI + WaveOpt-Estimator &0.90 &0.82   &46  \\
    NPI + WaveOpt-Estimator &0.94 &0.88   &28  \\
    \bottomrule 
  \end{tabularx}
  \label{tab:comparison2}
\end{table}
\vspace{-0.6cm}

\section{CONCLUSION}
\label{sec:majhead}

This paper presents a solution to improve the temporal constraints of Null-text Inversion (NTI) in the image editing process. Our newly introduced WaveOpt-Estimator effectively determines the endpoint $t^*$ of text optimization by analyzing the frequency characteristics of images using wavelet transform analysis. Notably, the WaveOpt-Estimator can be seamlessly applied to image editing models based on DDIM sampling, ensuring the high quality of the edited image. This research proposes a new direction that pursues both efficiency and quality in the field of image editing.

\vfill\pagebreak



\bibliographystyle{IEEEbib}
\bibliography{strings,refs}

\end{document}